\documentclass{article}
\usepackage{spconf,amsmath,graphicx,hyperref}
\usepackage{booktabs}
\usepackage[table]{xcolor}
\usepackage{amssymb}
\definecolor{bestcolor}{RGB}{252,228,226}    
\definecolor{secondcolor}{RGB}{223,237,248}  

\makeatletter
\renewcommand\section{\@startsection {section}{1}{\z@}%
                             {-7pt plus -2pt minus -1pt}
                             {4.5pt plus 1pt minus 1pt}
                             {\large\bf\A描写 \centering}} 

\renewcommand\subsection{\@startsection{subsection}{2}{\z@}%
                               {-5pt plus -1.5pt minus -1pt}
                               {3.5pt plus 1pt minus 1pt}
                               {\bf\A描写}} 
\makeatletter
\title{RouTex: Multi-Reference 3D Texture Generation by Reprogramming Rotary Position Encoding}
\name{Chengyang Li, Yang Liu, Jiale Zhang, Qiang Rao, Wei Zhang, Taicheng Huang\textsuperscript{*}
\thanks{\textsuperscript{*}Corresponding author: huangtaicheng.1@jd.com}}

\address{JD.com, Beijing, China}
\begin{document}
%
\maketitle
%

\begin{abstract}
While single-image 3D texture generation has advanced significantly, conditioning texture synthesis on multi-view reference images remains largely under-explored, despite its necessity for resolving surface ambiguity in unobserved regions.
In this paper, we propose RouTex, a novel multi-reference 3D texture generation framework based on in-context Diffusion Transformers (DiT).
Our key insight is that native 3D Rotary Position Encoding (RoPE) inherently possesses spatial addressing capabilities.
We introduce \textbf{Positional Routing (PR)} to reprogram 3D RoPE along spatial and semantic dimensions, establishing explicit reference-to-view correspondence without introducing additional parameters.
To execute this routing efficiently, \textbf{Dual-Granularity Attention (DGA)} combines per-view local attention for fine texture fidelity with global joint attention over deduplicated references for cross-view consistency.
Extensive experiments demonstrate that RouTex achieves new state-of-the-art performance in multi-reference 3D texture generation.
\end{abstract}

\section{Introduction}
\label{sec:intro}

High-quality 3D assets with rich surface textures are essential for e-commerce, gaming, and augmented reality, typically acquired via \emph{reconstruction} or \emph{generation}.
While reconstruction methods~\cite{kerbl3Dgaussians,mildenhall2020nerf,pan2024glomap} require dense, precisely calibrated multi-view captures, generative methods~\cite{lai2026natex,lumitex2026,xiang2025structured} synthesize 3D content from a text prompt or a single image.
However, such inputs of generation methods provide insufficient guidance for texture synthesis: text lacks fine-grained appearance control, while a single image leaves most surface areas unobserved.
Worse, because the mapping from visible to occluded regions is one-to-many (e.g., a mug with a bird on the front may have a plain or a floral back), synthesized back-view textures must rely entirely on learned priors, often producing hallucinated or inaccurate details.
Conditioning texture generation on multiple reference images is therefore crucial, as each additional view provides explicit appearance supervision for different surfaces of the object.

A straightforward approach to multi-reference conditioning is the cyclic injection scheme from TRELLIS~\cite{xiang2025structured}, which applies one reference per denoising step.
However, due to weak 3D priors and limited texture representation capabilities, TRELLIS recovers only coarse base textures, leading to severe distortion on fine details like text and logos.
We thus turn to the prevailing \textbf{Multi-View Diffusion (MVD)} framework~\cite{huang2025mv,seed2025seed3d,shi2023MVDream}.
Directly transplanting cyclic injection into a MVD framework fails, as it lacks 3D structural priors to auto-align mesh surfaces with reference images via cross-attention.
Consequently, the central challenge becomes: \emph{how to establish explicit correspondences between reference images and target views within a pure 2D framework.}



To address this challenge, our key observation is that modern in-context Diffusion Transformers (DiT)~\cite{ho2020denoising, Peebles2022DiT}, such as Flux-Kontext~\cite{labs2025flux1kontextflowmatching}, inherently possess spatial addressing capabilities via 3D Rotary Position Encoding (RoPE), where axis-0 of the position triplet distinguishes different images, while the two spatial axes encode 2D layout.
We therefore propose to \emph{reprogram} this native mechanism, without adding parameters, to inject explicit reference-to-view correspondence into generation.

Concretely, we introduce two complementary strategies to realize this design.
First, \textbf{Positional Routing (PR)} partitions camera yaws into canonical azimuth regions to group target views and reference images, and reprograms 3D RoPE to align each reference's spatial coordinates with its corresponding target view, enabling self-attention to automatically route appearance guidance (Sec.~\ref{ssec:routing}).
Second, \textbf{Dual-Granularity Attention (DGA)} leverages the native double-stream and single-stream blocks of Flux-Kontext to execute this routing efficiently, combining per-view local attention for fine texture fidelity with global joint attention over deduplicated references for cross-view consistency (Sec.~\ref{ssec:dga}).
Crucially, because correspondence is governed purely by position encodings across discrete regions, RouTex requires only coarse azimuth categories rather than precise camera calibration, substantially relaxing capture requirements while achieving state-of-the-art performance.

\section{Method}
\label{sec:method}

\begin{figure*}[t]
  \centering
  \includegraphics[width=\textwidth]{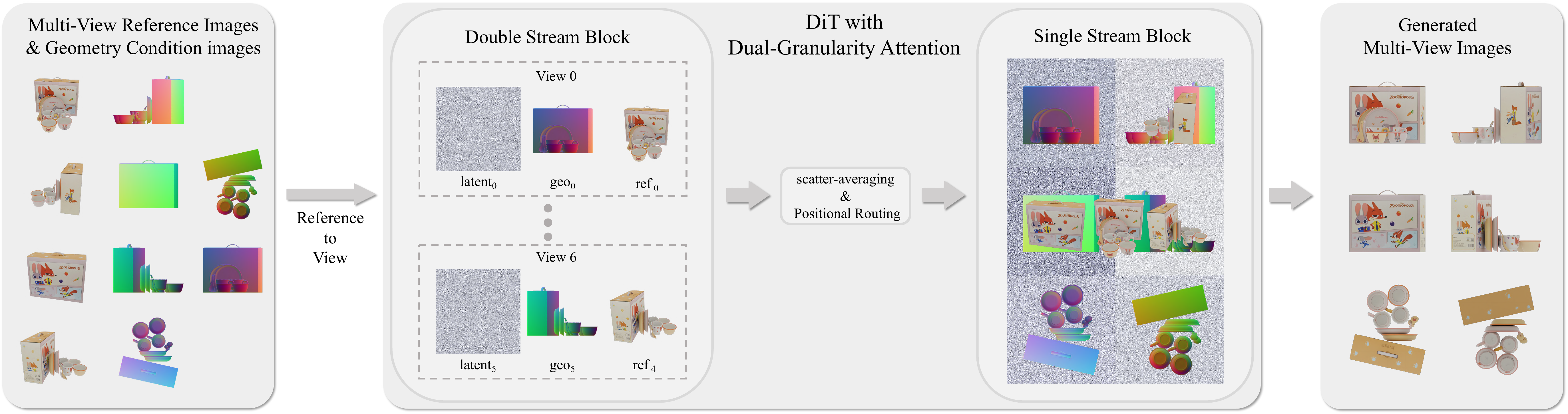}
  \caption{Overview of RouTex. Given multi reference images from different azimuth regions and geometry conditions, Positional Routing assigns fractional semantic codes and aligns each reference to its target view coordinates. Double-stream blocks perform per-view local attention, then single-stream blocks perform global joint attention with deduplicated references.}
  \label{fig:pipeline}
\end{figure*}

\subsection{Preliminaries}
\label{ssec:prelim}

\textbf{Flux-Kontext} is an in-context image-editing Diffusion Transformer (DiT) that concatenates reference and target tokens into a unified sequence, processed by $N_d$ double-stream blocks followed by $N_s$ single-stream blocks. It encodes positions with 3D Rotary Position Encoding (RoPE) using triplets $(d,y,x)$, where the semantic axis $d$ distinguishes different input images and the spatial axes $(y,x)$ specify the 2D layout.

\textbf{Problem Setup.}
Following multi-view diffusion protocols~\cite{lumitex2026, huang2025mv}, we arrange six orthographic target views into a $3{\times}2$ grid, as illustrated in Fig.~\ref{fig:pipeline}.
Given a 3D mesh, text prompt $t$, multi-view geometry conditions, and $K$ reference images $\text{ref}=\{\text{ref}_k\}_{k=0}^{K-1}$ ($1 \le K \le 4$), RouTex aims to generate high-fidelity textures consistent across all views. For each target view $v$, we denote its latent representation and geometry condition as $\text{latent}_v$ and $\text{geo}_v$ respectively. 

\subsection{Positional Routing}
\label{ssec:routing}

To establish explicit reference-to-view correspondence within a pure 2D framework, we propose \textbf{Positional Routing (PR)}, which reprograms the native 3D RoPE triplets $(d,y,x)$ along both spatial and semantic dimensions without adding parameters or modifying the architecture.

\textbf{Coordinate-Space Routing.}
We divide the $360^\circ$ camera yaw into four canonical azimuth regions $r \in \{0,1,2,3\}$ (front, right, back, and left). Each reference and target view is assigned to a region according to its camera yaw, except the two polar views (top and bottom with pitch $\pm90^\circ$), which are assigned to region 0 by default.

Let a single VAE latent have spatial size $h\times w$. For a token with local coordinate $(y,x)$ in the view at row $i$ and column $j$ of the $3{\times}2$ canvas, its global RoPE coordinate is
\begin{equation}
y' = y + ih,\qquad
x' = x + jw,
\label{eq:canvas_coord}
\end{equation}
where $i\in\{0,1,2\}$ and $j\in\{0,1\}$. 

For a reference in region $r\neq0$, we align its spatial coordinates with those of the first target view $v^*$ in the same region:
\begin{equation}
(y'_{\text{ref}_r},x'_{\text{ref}_r})
=
(y'_{v^*},x'_{v^*}).
\label{eq:routing}
\end{equation}
The region-0 reference is instead positioned at the center of the $3{\times}2$ canvas to broadcast its front-view appearance to all target views, as shown in Fig.~\ref{fig:pipeline}. Thus, relevant reference and target tokens share aligned spatial positions in the RoPE space, enabling parameter-free appearance routing.

\textbf{Fractional Semantic Addressing.}
To further distinguish token types, views, and reference regions, we reprogram the semantic axis $d$ as follows:
\begin{equation}
d = \begin{cases}
1 + v/V, & \text{for target latent of view } v,\\
2 + v/V, & \text{for geometry condition of view } v,\\
3 + r/R, & \text{for reference image in region } r,
\end{cases}
\label{eq:axis0}
\end{equation}
where $V=30$ is the candidate view pool size and $R=4$ is the number of azimuth regions. Six target views are sampled per iteration while retaining their intrinsic indices $v\in\{0,\dots,V-1\}$, allowing the fractional offsets $v/V$ and $r/R$ to distinguish views and reference regions.

\subsection{Dual-Granularity Attention}
\label{ssec:dga}

Positional Routing specifies \emph{which} reference is associated with each target view. However, six target views outnumber the available references ($K\le4$) and directly duplicating references is computationally inefficient, we therefore introduce \textbf{Dual-Granularity Attention (DGA)} to exploit the native two-stage architecture of Flux-Kontext for local texture transfer and global cross-view consistency.

\textbf{Double-Stream Blocks (Per-View Local Attention).}
During the double-stream stage, we isolate each target view into an independent sequence:
$[\text{latent}_v; \text{geo}_v; \text{ref}_{m(v)}]$,
where $m(v)$ denotes the matched reference for view $v$. 
By reshaping the batch dimension from $B$ to $B \times 6$, attention is restricted to each view-reference pair, reducing sequence length and enabling efficient fine-grained texture transfer.

\textbf{Single-Stream Blocks (Global Joint Attention).}
Before the single-stream stage, reference features from different view-reference pairs are merged into $K$ unique representations $\mathbf{h}_{0:K-1}$ via scatter-averaging.
The input sequence is then reconstructed as
$[t; \text{latents}; \text{geos}; \mathbf{h}_{0:K-1}]$,
allowing joint attention across all target views and unique references. This global interaction promotes cross-view appearance consistency while avoiding redundant reference computation.

Together, the double-stream blocks preserve local texture details at lower compute cost, while the single-stream blocks enforce global cross-view coherence to get high-fidelity texture.

\subsection{Training}
\label{ssec:training}

We train the entire model using the vanilla flow-matching~\cite{lipman2022flow} loss.
The number of reference images $K$ is randomly sampled from $\{1,2,3,4\}$, strictly requiring at least one reference from region 0. 
Our view pool consists of 30 candidate viewpoints, comprising six canonical orthographic views (front, back, left, right, top, and bottom) and 24 surrounding views. 
At each iteration, six target views are randomly sampled from this pool while ensuring full coverage of all four azimuth regions. 
For geometry conditioning, the surface normal map and canonical coordinate map (CCM) are averaged into a single composite condition map per view.

\section{Experiments}
\label{sec:exps}

\subsection{Implementation Details}
\label{ssec:impl}

Our training dataset comprises approximately 200K 3D assets curated from public datasets (Objaverse~\cite{deitke2023objaverse}, ABO~\cite{collins2022abo}, 3D-FUTURE~\cite{fu20213d}, TexVerse~\cite{zhang2025texverse}) alongside proprietary high-quality handcrafted models.
For each asset, we render 24 reference views at $2048{\times}2048$ resolution following the camera setup of TRELLIS~\cite{xiang2025structured}, and 30 orthographic target views at $1024{\times}1024$ as ground truth.
We train RouTex on 8 NVIDIA B200 GPUs with a batch size of 64 for 5.5K iterations (${\sim}32$ hours) at $512{\times}512$ resolution.
During inference, we scale the resolution to $768{\times}768$, generating six target views in approximately 40 seconds over 15 denoising steps without classifier-free guidance.
FlashAttention V4~\cite{zadouri2026flashattention} is employed across both training and inference to optimize runtime efficiency.

\subsection{Experimental Settings}
\label{ssec:setup}

\textbf{Baselines and Metrics.}
We compare RouTex against five state-of-the-art baselines: two multi-reference methods (TRELLIS~\cite{xiang2025structured} and TRELLIS.2~\cite{xiang2026native}) and three single-reference methods (Hunyuan3D-2.1~\cite{hunyuan3d2025hunyuan3d}, MV-Adapter~\cite{huang2025mv}, and LumiTex~\cite{lumitex2026}).
Specifically, multi-image input is natively supported in the official codebase of TRELLIS, whereas for TRELLIS.2, we adapt it using the same cyclic injection mechanism as TRELLIS for fair evaluation. Besides, all other baselines run under their default single-view setups.
To isolate texture generation quality from geometric variations, all methods texture the same meshes reconstructed by Hunyuan3D-2.1 and decimated to 50K faces.
Outputs are evaluated by baking synthesized textures back onto these shared meshes, measuring quality via FID~\cite{heusel2017gans}, KID~\cite{binkowski2018demystifying}, CLIP-I~\cite{radford2021learning}, and LPIPS~\cite{zhang2018unreasonable}.

\textbf{Evaluation Dataset.}
We curate an independent evaluation set of 475 high-quality 3D assets, disjoint from the training data.
For each asset, we render 12 evaluation views at a $15^\circ$ elevation angle, uniformly sampled every $30^\circ$ in azimuth starting from $0^\circ$.
To eliminate illumination bias arising from differences between PBR-based and shaded texture pipelines among baselines, all evaluation renderings are performed under uniform ambient lighting.
RouTex and other multi-reference baselines take four canonical orthogonal views (front, back, left, right) as condition images, whereas single-reference baselines use the front view.

\subsection{Quantitative Results}
\label{ssec:main}

\begin{table}[t]
\centering
\renewcommand{\arraystretch}{1.2}
\begin{tabular}{lcccc}
\toprule
Method & FID$\downarrow$ & KID$\downarrow$ & CLIP-I$\uparrow$ & LPIPS$\downarrow$ \\
\midrule
TRELLIS        & 18.83 & 28.49 & 0.9149 & 0.1727 \\
TRELLIS.2     & \cellcolor{secondcolor}12.93 & \cellcolor{bestcolor}4.53 & 0.9246 & \cellcolor{secondcolor}0.1656 \\
Hunyuan3D-2.1 & 15.52 & 10.34 & 0.9242 & 0.1703 \\
MV-Adapter  & 15.34 & 12.04 & 0.9217 & 0.1710 \\
LumiTex        & 17.64 & 23.82 & \cellcolor{secondcolor}0.9258 & 0.1666 \\
Ours                           & \cellcolor{bestcolor}12.06 & \cellcolor{secondcolor}8.66 & \cellcolor{bestcolor}0.9423 & \cellcolor{bestcolor}0.1519 \\
\bottomrule
\end{tabular}
\caption{Quantitative comparison with state-of-the-art methods. \colorbox{bestcolor}{Red} and \colorbox{secondcolor}{blue} highlight the best and second-best results, respectively.}
\label{tab:comparison}
\end{table}

As shown in Table~\ref{tab:comparison}, RouTex outperforms all baselines.
Single-reference methods see only the front view, failing to hypothesize plausible textures for unobserved surfaces.
Multi-reference methods suffer from weak 3D priors and limited texture representation capacity, while their iterative injection scheme lacks explicit reference-to-view correspondence, resulting in severe texture distortion.
In contrast, RouTex explicitly routes multi-reference texture within a strong generative framework, producing high-quality textures that remain plausible and consistent across all views.

\subsection{Qualitative Results}
\label{ssec:qualitative}

\begin{figure*}[t]
  \centering
  \includegraphics[width=\textwidth]{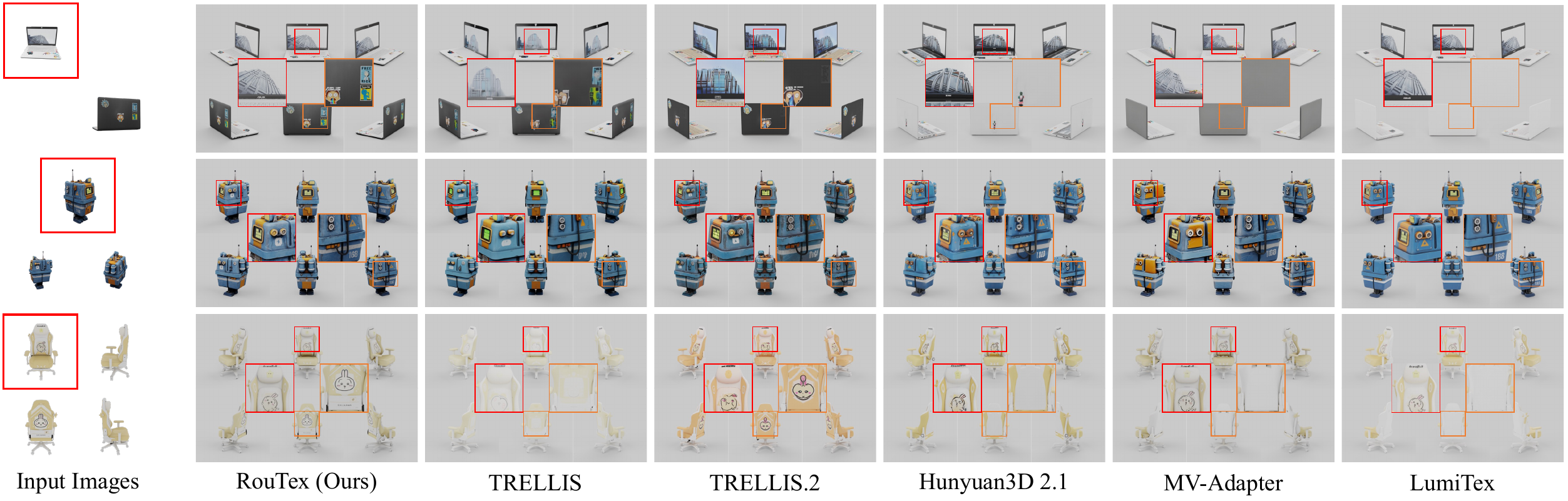}
  \caption{Qualitative comparison with open-source methods. Orange and red boxes highlight zoomed-in regions for detail inspection. In the input column, red-bordered images indicate the input of single-image methods.}
  \label{fig:open_source_compare}
\end{figure*}

\textbf{Open-Source Methods.}
Fig.~\ref{fig:open_source_compare} compares RouTex with open-source state-of-the-art baselines using 2 to 4 reference images, where RouTex faithfully reproduces complex textures across all input views.
Single-reference methods receive no visual guidance for unobserved regions, often generating flat, uninformative surfaces. Even within observed areas, they fail to reconstruct fine-grained appearance details.
While multi-reference methods capture basic color distributions, their limited 3D texture representations and iterative injection scheme lack explicit reference-to-view correspondence, leaving fine patterns and text noticeably blurred or garbled.

\begin{figure}[t]
  \centering
  \includegraphics[width=\columnwidth]{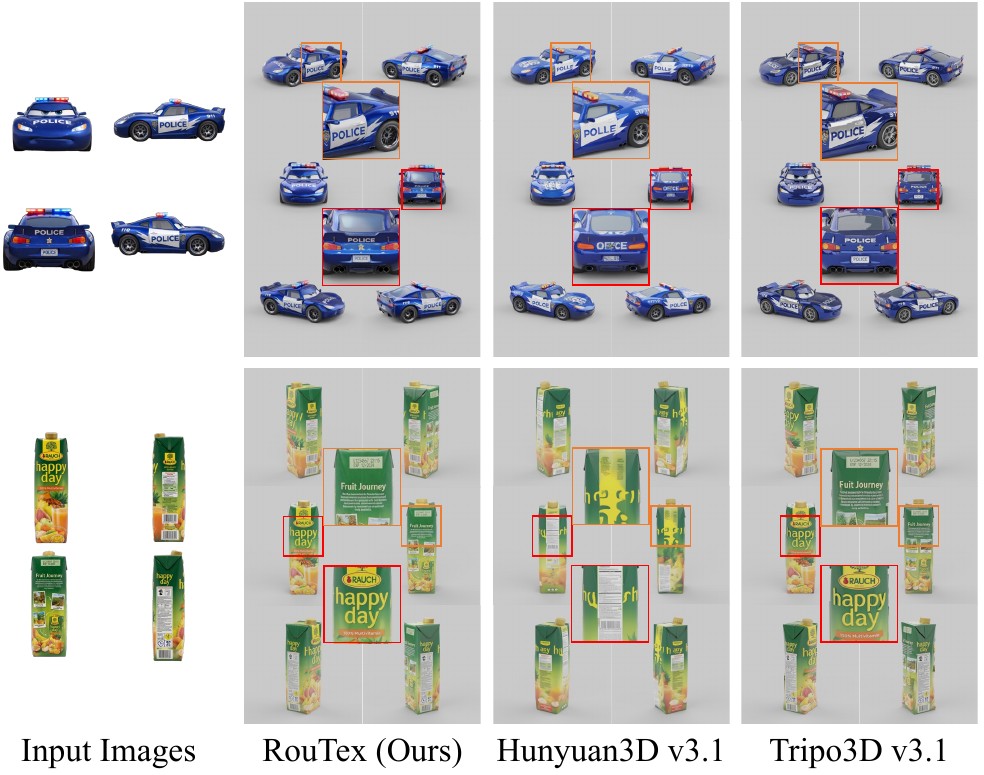}
  \caption{Qualitative comparison with commercial methods. Orange and red boxes highlight zoomed-in regions for detail inspection.}
  \label{fig:commercial_compare}
\end{figure}
\textbf{Commercial Methods.}
We further compare against two commercial models that support multi-image 3D generation.
To reduce the resolution gap for a fairer comparison, we retrain our model at $1024{\times}1024$ resolution.
As shown in Fig.~\ref{fig:commercial_compare}, Hunyuan3D v3.1 produces chaotic textures with severe cross-view inconsistencies.
Tripo3D v3.1 generates cleaner textures overall but struggles with spatial precision and fine details. As highlighted in the zoomed-in insets, the painting in the orange box (column~4, row~2) is misaligned relative to the reference image, and the text in the red box is less crisp than ours.

\subsection{Ablation Study}
\label{ssec:ablation}

\begin{table}[t]
\centering
\renewcommand{\arraystretch}{1.2}
\begin{tabular}{lcccc}
\toprule
Method & FID$\downarrow$ & KID$\downarrow$ & CLIP-I$\uparrow$ & LPIPS$\downarrow$ \\
\midrule
CSIC   & 20.08   & 27.49   & 0.9122    & 0.1700    \\
w/o PR \& DGA   & 14.61   & 13.13   & 0.9280    & 0.1619    \\
w/o PR          & 12.49 & 9.63  & 0.9395 & 0.1541 \\
w/o DGA         & \cellcolor{secondcolor}12.11 & \cellcolor{bestcolor}8.36 & \cellcolor{secondcolor}0.9413 & \cellcolor{secondcolor}0.1523 \\
Full Model     & \cellcolor{bestcolor}12.06 & \cellcolor{secondcolor}8.66 & \cellcolor{bestcolor}0.9423 & \cellcolor{bestcolor}0.1519 \\
\bottomrule
\end{tabular}
\caption{ Quantitative results of ablation study. \colorbox{bestcolor}{Red} and \colorbox{secondcolor}{blue} highlight the best and second-best results, respectively.}
\label{tab:ablation}
\end{table}

\begin{figure}[t]
  \centering
  \includegraphics[width=\columnwidth]{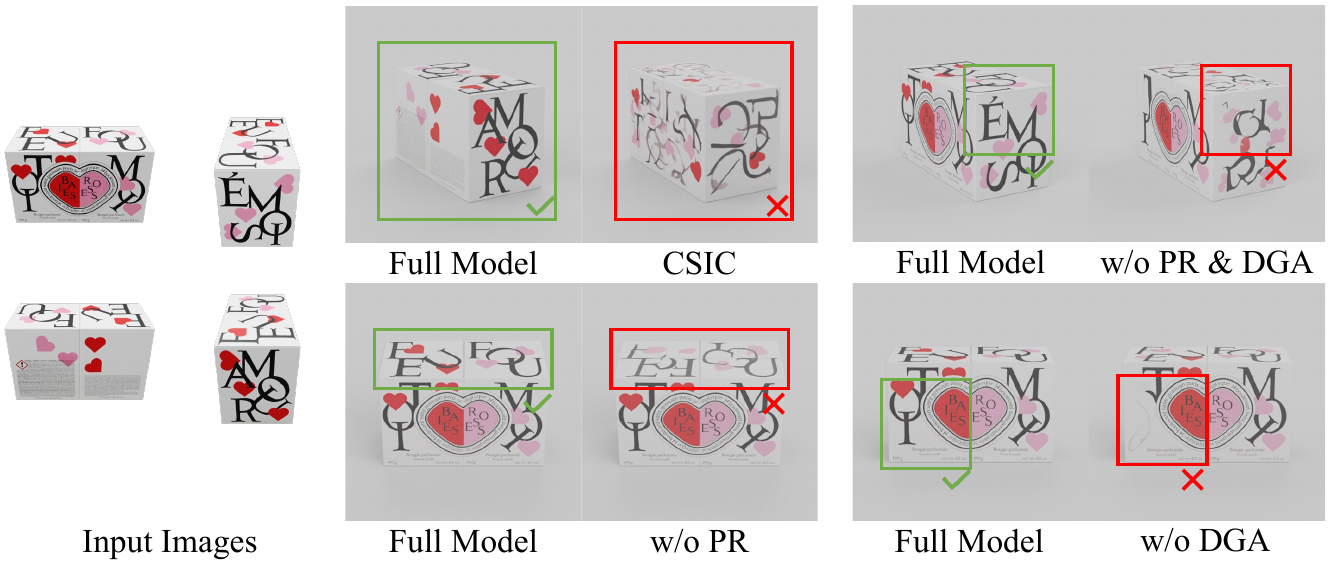}
  \caption{Qualitative ablation results. Green and red boxes highlight successful recoveries by the Full Model and failures of the ablated variants, respectively.}
  \label{fig:ablation}
\end{figure}

We validate the necessity of each key component through four ablation variants: 
(1)~\textbf{CSIC} (Cyclic Single-Image Conditioning) adopts TRELLIS's injection strategy with the single reference MVD framework;
(2)~\textbf{w/o PR} removes Positional Routing, assigning identical RoPE to all reference images in single-stream blocks while retaining per-view matching in double-stream blocks; 
(3)~\textbf{w/o DGA} processes all target views and references jointly across both double-stream and single-stream stages while preserving Positional Routing; 
(4)~\textbf{w/o PR \& DGA} disables both components.

As reported in Table~\ref{tab:ablation} and Fig.~\ref{fig:ablation}, removing any core component or replacing our multi-reference injection with TRELLIS's strategy leads to a performance degradation.
Furthermore, although removing DGA leads to only a slight quantitative drop, it remains essential for preventing localized texture artifacts and enabling a ${\sim}17\%$ inference speedup.



\section{Conclusion}
\label{sec:conclusion}
In this work, we present a multi-reference framework for high-quality 3D texture generation within a 2D multi-view diffusion framework. We introduce Positional Routing (PR) to establish explicit reference-to-view correspondence through reprogrammed 3D RoPE, and Dual-Granularity Attention (DGA) to efficiently combine local texture transfer with global cross-view consistency. With these components, RouTex achieves state-of-the-art performance in multi-reference 3D texture generation.

\bibliographystyle{IEEEbib}
\bibliography{refs}

\end{document}